\documentclass[letterpaper]{article} 
\usepackage{aaai24}  
\usepackage{times}  
\usepackage{helvet}  
\usepackage{courier}  
\usepackage[hyphens]{url}  
\usepackage{graphicx} 
\usepackage{natbib}  
\usepackage{caption} 
\usepackage{amsmath}
\usepackage{amsthm}
\usepackage{algorithm}
\usepackage{algorithmic}
\usepackage{multirow}
\usepackage{amsfonts}
\usepackage{amssymb}
\usepackage{color}

\usepackage{newfloat}
\usepackage{listings}
\DeclareCaptionStyle{ruled}{labelfont=normalfont,labelsep=colon,strut=off} 
\floatstyle{ruled}
\newfloat{listing}{tb}{lst}{}
\floatname{listing}{Listing}
\title{AltDiffusion: A Multilingual Text-to-Image Diffusion Model}
\author {
    Fulong Ye \footnotemark[1] \footnotemark[2] \textsuperscript{\rm 1} \textsuperscript{\rm 2}, 
    Guang liu \footnotemark[1] \textsuperscript{\rm 2},
    Xinya Wu \textsuperscript{\rm 2}, 
    Ledell Wu \textsuperscript{\rm 2}}
\affiliations {
    \textsuperscript{\rm 1} Beijing University of Posts and Telecommunications, Beijing, China \\
    \textsuperscript{\rm 2} Beijing Academy of Artificial Intelligence \\
    \texttt{fulong\_ye@bupt.edu.cn} \\
    \texttt{\{liuguang, yxwu, wuyu\}@baai.ac.cn}
}

\usepackage{bibentry}

\begin{document}

\maketitle

\renewcommand{\thefootnote}{\fnsymbol{footnote}} 
\footnotetext[1]{Equal contribution.}
\footnotetext[2]{Work done during internship with Beijing Academy of Artificial Intelligence.}

\renewcommand{\thefootnote}{\arabic{footnote}}

\begin{abstract}
Large Text-to-Image(T2I) diffusion models have shown a remarkable capability to produce photorealistic and diverse images based on text inputs. However, existing works only support limited language input, e.g., English, Chinese, and Japanese, leaving users beyond these languages underserved and blocking the global expansion of T2I models. Therefore, this paper presents AltDiffusion, a novel multilingual T2I diffusion model that supports eighteen different languages\footnote{Eighteen languages: English, Chinese, Japanese, Thai, Korean, Hindi, Ukrainian, Arabic, Turkish, Vietnamese, Polish, Dutch, Portuguese, Italian, Spanish, German, French, and Russian.}. Specifically, we first train a multilingual text encoder based on the knowledge distillation. Then we plug it into a pretrained English-only diffusion model and train the model with a two-stage schema to enhance the multilingual capability, including concept alignment and quality improvement stage on a large-scale multilingual dataset. Furthermore, we introduce a new benchmark, which includes Multilingual-General-18(MG-18) and Multilingual-Cultural-18(MC-18) datasets, to evaluate the capabilities of T2I diffusion models for generating high-quality images and capturing culture-specific concepts in different languages. Experimental results on both MG-18 and MC-18 demonstrate that AltDiffusion outperforms current state-of-the-art T2I models, e.g., Stable Diffusion in multilingual understanding, especially with respect to culture-specific concepts, while still having comparable capability for generating high-quality images. All source code and checkpoints could be found in https://github.com/superhero-7/AltDiffuson.

\end{abstract}

\begin{figure*}[htbp]
  \centering 
  \includegraphics[width=0.95\textwidth]{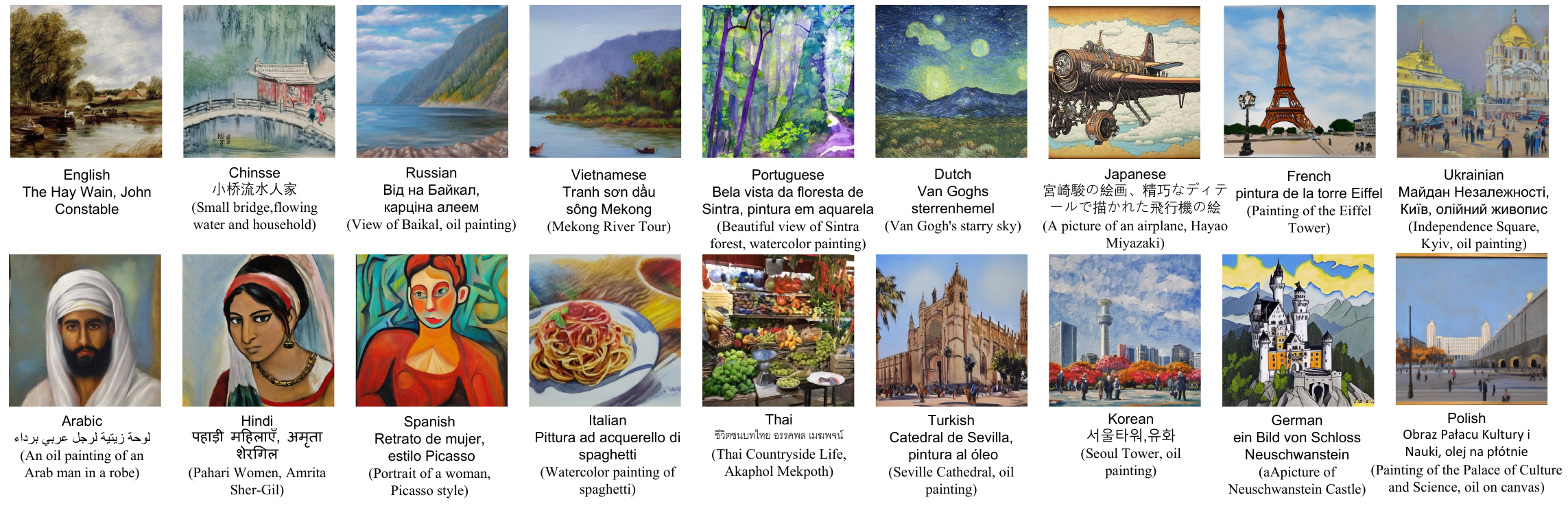}
  \caption{Images generated by AltDiffusion with prompts in various languages. We select prompts with culture-specific concepts in different languages to demenstrate the strong capability of multilingual T2I generation of AltDiffusion.}
  \label{fig: generated_exmaples}
\end{figure*}








\section{Introduction}
\label{sec:intro}



In recent years, there has been an emerging interest in large Text-to-Image(T2I) diffusion models, such as Stable Diffusion(SD)\cite{stable_diffusion}, Imagen\cite{imagen} and DALLE2\cite{dalle2}, due to their remarkable ability to produce photorealistic and diverse images based on text input. A limitation of these large T2I diffusion models is that they only support prompts in English, which is inconvenient for non-English users, e.g., Spanish or French. Non-English users usually utilize T2I diffusion models with the help of translation tools, which may lead to translation error and information loss, especially in some culture-specific concepts. For example, the name of the famous Chinese painter Baishi Qi may be translated as "white stone" in English. And the translation process is getting more complex when it comes to mixed language prompts. Intuitively, a T2I generative model uses native languages directly without additional translation steps have no such problems. Recently, some scholars have begun to develop multilingual T2I diffusion models. Taiyi-Bilingual\cite{wang2022fengshenbang}, ERNIE-ViLG 2.0\cite{feng2023ernie} are T2I bilingual models that support both Chinese and English. However, these T2I diffusion models are still limited by the scarcity of language varieties. 


To address the problem, we propose a novel multilingual T2I diffusion model, which is capable of processing eighteen languages that cover 46.94\% of the world's first-language speakers and 27.64\% of the world's second-language speakers, named AltDiffusion(AD), along with an efficient training approach. We first train a multilingual text encoder based on the knowledge distillation\cite{altclip} to enhance the language capability to support eighteen languages. Then, the parameters of the text encoder are frozen and plugged into a pretrained English-only diffusion model. Next, we propose a two-stage training schema to enhance the language capability of the diffusion model. In the first stage, we train the K, V matrix in cross-attention of the UNet on a multilingual dataset LAION 5B\cite{laion} to align the embedding space between the UNet and the text encoder. In the second stage, all parameters of the UNet are unfrozen to improve the quality of the generated images using LAION Aesthetics\cite{laion_aethetics}. In addition, a classifier-free guidance training technique is employed to further improve the generation quality. It is worth noting that our training approach possesses strong generality and can support any pretrained T2I diffusion models.


To evaluate our AD model, we introduce a benchmark including two evaluation datasets that focus on two aspects of multilingual generation, respectively. For general quality evaluation, we expand the data of XM-3600 by filtering high-quality image-text pairs from WIT and construct a high-quality dataset Multilinguale-General-18(\textbf{MG-18}) that includes 7,000 images per language to evaluate FID\cite{fid}, IS\cite{is}, and CLIP Sim\cite{clipscore}. For culture-specific concepts evaluation, we introduce Multilinguale-Cultural-18(\textbf{MC-18}), a culture-specific dataset with 50 text prompts per language about culture-specific concepts of different countries. The MC-18 is the first dataset about culture-specific concepts. This benchmark provides robust and comprehensive evaluation for multilingual T2I generation.

Our experimental results on MG-18 demonstrate that AD is the first multilingual T2I diffusion model that supports eighteen languages and outperforms other multilingual diffusion models, e.g. Taiyi\cite{taiyi} and Japanese SD models\cite{japanese_stable_diffusion}, in FID, IS, and CLIP Sim. In addition, AD surpasses translation-based SD in CLIP Sim of all languages and achieves comparable results in FID and IS, proving that AD is better than SD in multilingual understanding and can generate almost the same high-quality images as SD on general prompts. Experimental results on MC-18 show that AD beats translation-based SD in the culture-specific concepts in all languages.


Our contributions are as follows:



\begin{itemize}
\item{} We introduce AltDiffusion, a novel multilingual diffusion model that supports eighteen languages, which covers 46.94\% of the world's first-language speakers and 27.64\% of the world's second-language speakers.\footnote{https://en.wikipedia.org/wiki/List\_of\_languages\_by\_total\_num-ber\_of\_speakers}
\item{} We introduce a benchmark that includes two datasets for evaluating T2I generative model: a general quality evaluation dataset MG-18 and a culture-specific dataset MC-18. This benchmark provides robust and comprehensive evaluation for multilingual T2I generation.
\item{} AltDiffusion outperforms other multilingual diffusion models and performs better than a translation-based Stable Diffusion in multilingual understanding capability, especially in culture-specific concepts. 



\end{itemize}

\section{Related Work}

\begin{figure*}[htbp] 
\centering 
\includegraphics[width=1.0\textwidth]{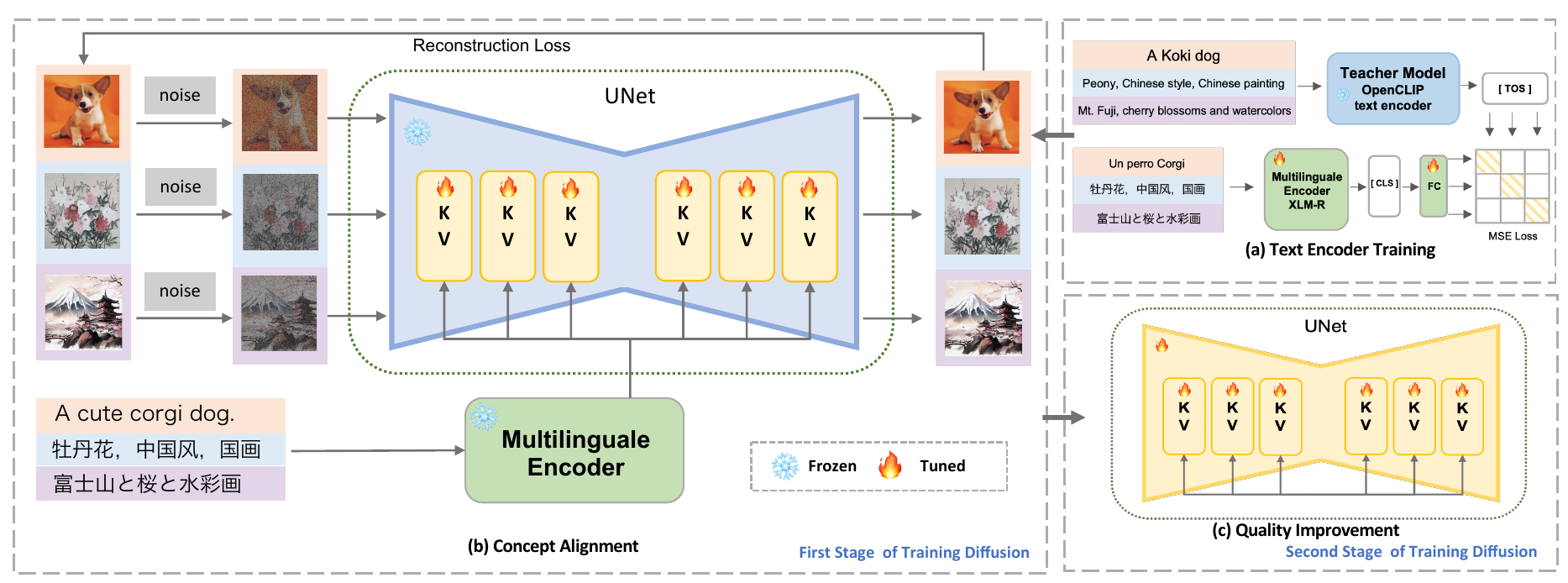} 
\caption{Illustration of the training approach. First, we train a multilingual text encoder. Then in the concept alignment stage, we only unfreeze the k and v parameters in cross-attention. In the quality improvement stage, all parameters of the UNet are unfrozen. Both stages are trained in 18 languages(Here only illustrate English, Chinese, and Japanese).}
\label{fig: model.pdf}
\end{figure*}


\noindent\textbf{Multilingual Text-to-image Generation}
 Recently, T2I diffusion models (\cite{stable_diffusion}, \cite{imagen}, \cite{dalle2}, \cite{vilg}, \cite{vilg2}, \cite{cogview}, \cite{cogview2}) achieve remarkable success in generating photorealistic and diverse images. Stable Diffusion(SD) is a prominent open-source framework with a considerable community. SD model consists of three parts: autoencoder is responsible for encoding images and decoding the pictures into and from the latent space; text encoder is accountable for encoding text prompts; Unet\cite{unet} is responsible for predicting noise based on the language embedding in latent space. Despite its strong generative capability, the fact that the SD model can only support English input still leads to limitations. Some works, such as CogView(\cite{cogview}, \cite{cogview2}) and ERNIE-ViLG(\cite{vilg}, \cite{vilg2}), start to explore the T2I diffusion model that can support multilingual text prompt. Only some studies try to extend the applicability of the SD beyond English to other languages, e.g., Taiyi\cite{taiyi} and Japanese SD~\cite{japanese_stable_diffusion}. However, these T2I generative models are still limited by the scarcity of language varieties. In this work, We are committed to constructing a multilingual T2I diffusion model which can serve most of the world's population. 
 
\noindent\textbf{Multilingual CLIP} CLIP shows a solid capability to provide a robust text-image representation in English. Recently, several works have tried to expand the language capabilities of CLIP to other languages. For instance, previous studies \cite{multilingual_1}, \cite{multilingual_2} attempt to create multilingual CLIP models. AltCLIP \cite{altclip} applies knowledge distillation techniques to develop a state-of-the-art multilingual CLIP model by leveraging the XLM-R multilingual model \cite{XLMR}. Following AltCLIP, we retrain the text encoder to align the penultimate hidden layer of OpenCLIP\cite{openclip} with the text encoder used in SD v2 to create a multilingual CLIP model. 

\noindent\textbf{Multilingual Image Caption Datasets} 
Image caption datasets are widely used for multimodal tasks, which are mainly accessible in English, e.g., Flickr30k\cite{Plummer_Wang_Cervantes_Caicedo_Hockenmaier_Lazebnik_2017}, MS COCO\cite{mscoco}. These monolingual datasets are limited by language-linguistic diversity. Thus some works have focused on image caption datasets in different languages. Multi30K\cite{Elliott_Frank_Sima’an_Specia_2016} is a dataset that supports German, while Wikimedia Commons\cite{Schamoni_Hitschler_Riezler_2018} can support German, French, and Russian. Recently, datasets that support multiple languages such as WIT\cite{srinivasan2021wit}, XM 3600\cite{thapliyal2022crossmodal} are proposed. WIT is collected by gathering diverse textual content linked to an image from Wikipedia, which has 37.6 million image-text pairs in 100+ languages. XM 3600 is a manually annotated dataset with 3,600 image-text in 36 languages. Based on XM3600 and WIT datasets, we build a dataset MG-18 to evaluate AD.




\section{Method}


The current Large T2I diffusion models usually consist of a text encoder and a UNet model. To train a multilingual T2I diffusion model, we first enhance the language capability of the text encoder and then align it with the UNet to enhance the language capability of UNet.
\begin{figure*}[htbp] 
\centering 
\includegraphics[width=1.0\textwidth]{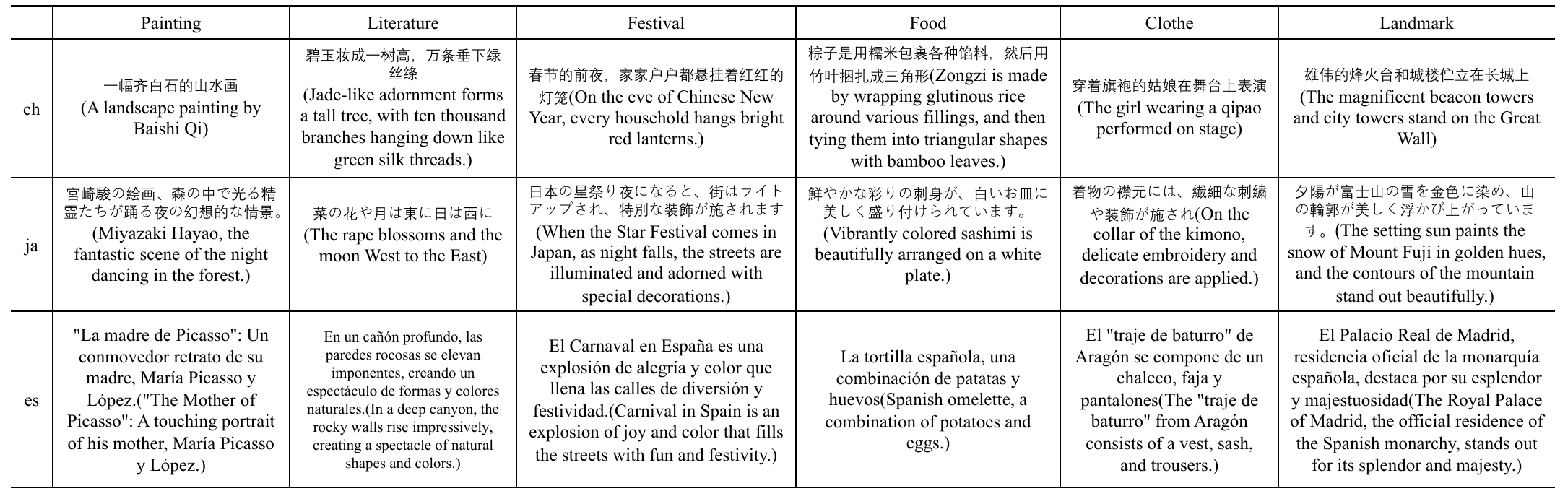} 
\caption{Data samples of MC-18 in Chinese(ch), Japanese(ja) and Spanish(es).}
\label{fig: MC-18}
\end{figure*}
\subsection{Enhance Language Capability of the Text Encoder}

Following AltCLIP\cite{altclip}, we retrain the text encoder to support 18 languages based on the knowledge distillation. As shown in Figure~\ref{fig: model.pdf}(a), the text encoder from OpenCLIP\cite{openclip} is the teacher model, and XLM-R\cite{XLMR} is the student model. Given parallel prompts ($text_{english}$, $text_{other language}$), the $text_{english}$ input is fed into the teacher model, and the $text_{other language}$ input is fed into the student model. We minimize the Mean Squared Error(MSE) between $[TOS]$ embedding of the teacher model and $[CLS]$ embedding of the student model. A fully connected network maps the outputs of XLM-R and the OpenCLIP text encoder penultimate layer to the same dimensionality. After training, we obtain a multilingual text encoder whose embedding space is close to the original OpenCLIP text encoder. 

\subsection{Enhance Language Capability of the UNet}

After training the text encoder, the parameters of the text encoder are frozen and plugged into an off-the-shelf pretrained English-only diffusion model(here, we use SD, but our method can be extended to other diffusion models with a text encoder). Then we use two-stage training schema, including concept alignment and quality improvement stage, to transform the English-only diffusion model into a multilingual one.

\textbf{Concept Alignment Stage}
This stage aims to re-establish the relationship between the text and the images by aligning the embedding space between the text encoder and UNet. The training dataset is LAION\cite{laion}, which is a large-scale dataset with a multilingual corpus(detailed introduction is in Dataset\ref{Dataset} section). Preliminary analysis\cite{erasing} reveals that the cross-attention mechanism of the diffusion model plays a crucial role in matching the text and images. Therefore, as illustrated in Figure~\ref{fig: model.pdf}(b), we freeze the multilingual text encoder, the autoencoder, and most of the parameters of the UNet, then train the K, V matrix of the cross-attention module using the denoising diffusion objective\cite{ddpm}:
\begin{equation}
\mathcal{L}=\mathbb{E}_{\varepsilon(x), t, c, \epsilon \sim N(0,1)}\left[\left\|\epsilon-\epsilon_O\left(z_t, c, t\right)\right\|_2^2\right]
\label{eq: denoising_loss}
\end{equation}
where $t \sim \text { Uniform }[1, T]$, $z_t$ is a noisy version of latent embedding $z$ of input image $x$ (i.e. $z=\mathcal{E}(x)$), obtained using $\epsilon \sim \mathcal{N}(0, \mathbf{I})$. $\theta$ is the parameter of UNet to predict the noisy $\epsilon_\theta\left(z_t, c, t\right)$ condition on $z_t$, multilingual text condition embedding $c$ and $t$. 

Previous observation\cite{stable_diffusion} indicates that reducing the dimensions of images from 512$\times$512 to 256$\times$256 results in minimal damage to semantic information, only eliminating some imperceptible details. In line with our objective of aligning the semantic information between modalities, we utilize the lower image resolution of 256$\times$256 during this stage to facilitate faster training while minimizing computational costs.

\textbf{Quality Improvement Stage}
In the second stage, as illustrated in Figure~\ref{fig: model.pdf}(c), we apply a continuous learning strategy by loading the first stage checkpoint and subsequently fine-tuning all the parameters of the UNet using the same objective function as defined in Equation~\ref{eq: denoising_loss}. We train our model on high-quality multilingual aesthetic datasets with a resolution of 512x512 to improve the generative quality. Furthermore, we also drop 10\% of text inputs following the setting as SD\cite{stable_diffusion} for classifier-free guidance\cite{cfg}, which is helpful when calculating the final noisy score during inference. $\tilde{\epsilon}_\theta\left(z_t, c, t\right)$ is obtained by a combination of conditioned score $\epsilon_\theta\left(z_t, c, t\right)$ and unconditioned score $\epsilon_\theta\left(z_t, t\right)$, which is formalized in the following equation:

\begin{equation}
\tilde{\epsilon}_\theta\left(z_t, c, t\right)=\epsilon_\theta\left(z_t, t\right)+\alpha\left(\epsilon_\theta\left(z_t, c, t\right)-\epsilon_\theta\left(z_t, t\right)\right)
\end{equation}

\noindent where $\alpha>1$ is a scale weight of condition. With the completion of the second stage, we finally obtain a multilingual T2I diffusion model that meet the needs of users across different linguistic backgrounds.


\section{Dataset \label{Dataset}}
\subsection{Training Data}
All the image-text pairs we use to train AD come from LAION \cite{laion}. The details of these image-text pairs are as follows:

\textbf{LAION 5B} LAION 5B includes three sub-datasets: LAION2B-en, LAION2B-multi and LAION1B-nolang. LAION2B-en contains 2.32 billion image-text pairs in English. LAION2B-multi contains 2.26 billions image-text pairs and the text comes from 100+ languages beyond English. In the first training stage, we filter 1.8 billions data in eighteen languages from LAION2B-multi and combine it with LAION2B-en.

\textbf{LAION Aesthetics} LAION Aesthetics contains several collections of subsets from LAION 5B with high quality. An Aesthetics Predictor is trained using LAION to predict the aesthetics score of images on a scale of 1 to 10, with higher aesthetics score being better. Then the Aesthetics Predictor is used for filtering the data. To conduct the second training stage, we filter eighteen languages from the LAION Aesthetics and the LAION Aesthetics V1-multi dataset with the predicted aesthetics score higher than seven.
\subsection{Evaluation Benchmark}
To evaluate the capability of AD to generate images and capture culture-specific concepts of different languages, we introduce two datasets: Multilingual-General-18(MG-18) for generation quality evaluation and Multilingual-Cultural-18(MC-18) for culture-specific concepts evaluation.

\textbf{Multilinguale-General-18(MG-18)}
We construct a large and high-quality dataset MG-18 which contains 7,000 image-text pairs in 18 languages, by expanding XM 3600 with high-quality images from WIT in two steps. In the first step, we use an Optical Character Recognition\cite{ocr} system to filter out images with more than five words, considering images with excessive text tend to be document images, which are unsuitable for evaluating the generation capabilities of the T2I model. Next, we use AltCLIP to calculate the similarity score(CLIP Sim) between the image and the caption, and then keep those with a score higher than 0.2.

\textbf{Multilinguale-Cultural-18(MC-18)}
One of the important capabilities of multilingual T2I models is to understand culture-specific concepts of different languages. To evaluate this, we construct MC-18, a dataset that contains culture-specific concepts in painting, literature, festival, food, clothe, and landmark. First, we select the representatives of each language in the above six aspects. Then we use ChatGPT to generate prompts and ask the crowdsourcing personnels to select suitable prompts. We create 50 prompts for each of the 18 languages. Some samples of MC-18 are shown in Figure~\ref{fig: MC-18}

\begin{figure*}[htbp] 
\centering 
\includegraphics[width=1\textwidth]{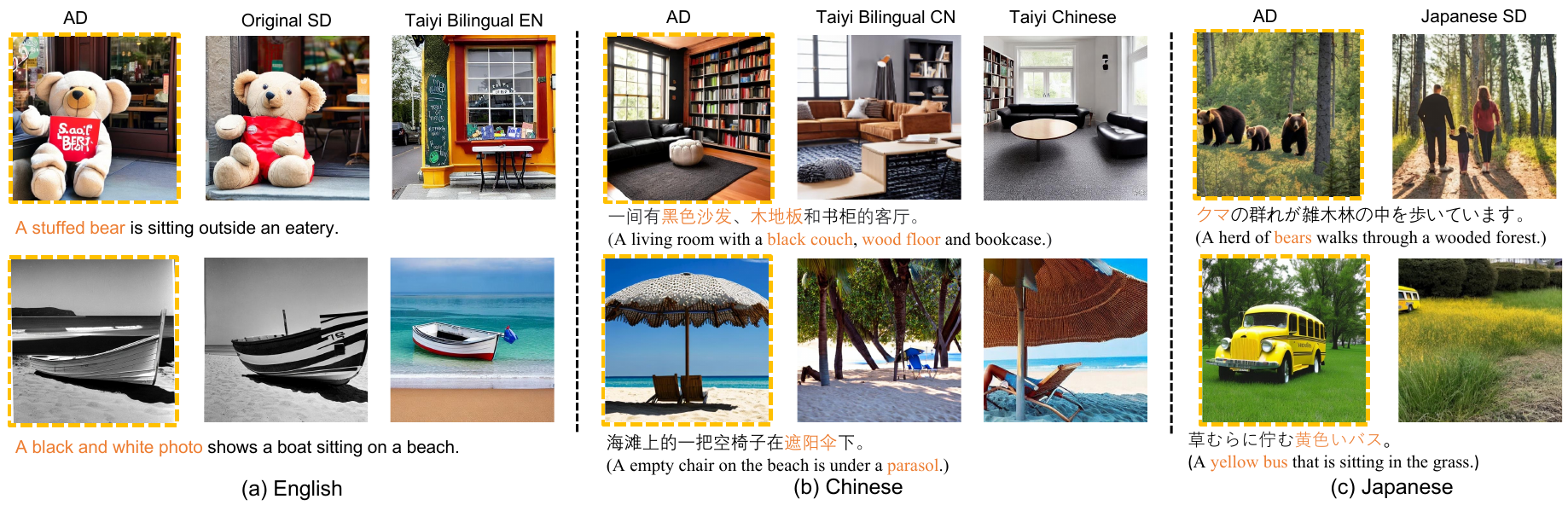} 
\caption{Comparison of generated results with original English SD and other multilingual diffusion models on MG-18. Results of AD are framed with yellow dashed boxes. Obvious differences in generation are highlighted in orange in the prompts.}
\label{fig: compared_results.pdf}
\end{figure*}

\section{Experiments}



\subsection{Implement Details \label{sec: implement_details}}
The optimizer is AdamW\cite{adamw}. The learning rate is 1e-4, with 10,000 warmup steps on 64 NVIDIA A100-SXM4-40GB GPUs.

Follow AltCLIP\cite{altclip}, we use knowledge distillation to retrain the multilingual text encoder. Through the training process, the text encoder remains frozen. In addition, we adopt a continuous learning strategy for model training. In the concept align stage, we use the SD v2.1 512-base-ema\footnote{https://huggingface.co/stabilityai/stable-diffusion-2-1-base} checkpoint to initialize all parameters except the text encoder, with a batch size of 3,072 and a resolution of 256x256. The training process on LAION2B-en and LAION2B-multi for 330,000 steps takes approximately eight days. In the quality improvement stage, the training starts at the 330,000-step checkpoint, with a batch size of 3,840 on LAION Aesthetics V1-en and V1-multi and 270,000-steps with a resolution of 512x512, which takes around seven days. After that, a new round of training continues from the 270,000-step checkpoint for another 150,000 steps, with 10\% of the text randomly discarded for classifier-free guidance learning, taking approximately four days. The teacher model using in knowledge distillation is OpenCLIP ViT-H-14\footnote{https://huggingface.co/laion/CLIP-ViT-H-14-laion2B-s32B-b79K}. We also use Xformer and Efficient Attention to save memory use and speed up training. The decay of EMA is 0.9999.

\subsection{Results on MG-18}

\begin{table}[htbp]
\centering
\resizebox{\columnwidth}{!}{
\begin{tabular}{c|ccc|ccc}
\hline
\multirow{2}{*}{} & \multicolumn{3}{c|}{AltDiffusion(AD)}                                  & \multicolumn{3}{l}{Stable Diffusion(SD)}                               \\ \cline{2-7} 
                  & \multicolumn{1}{c|}{FID}   & \multicolumn{1}{c|}{IS}    & CLIP Sim & \multicolumn{1}{c|}{FID}   & \multicolumn{1}{c|}{IS}    & CLIP Sim \\ \hline
English & \multicolumn{1}{l|}{19.02} & \multicolumn{1}{l|}{\textbf{27.45}} & \textbf{0.324} & \multicolumn{1}{l|}{18.02} & \multicolumn{1}{l|}{26.97} & \multicolumn{1}{c}{0.291} \\
Chinese           & \multicolumn{1}{l|}{20.32} & \multicolumn{1}{l|}{\textbf{29.46}} & \multicolumn{1}{c|}{\textbf{0.350}}    & \multicolumn{1}{l|}{\textbf{18.51}} & \multicolumn{1}{l|}{28.30} & 0.317    \\
Japanese          & \multicolumn{1}{l|}{18.90} & \multicolumn{1}{l|}{28.13} & \multicolumn{1}{c|}{\textbf{0.356}}    & \multicolumn{1}{l|}{\textbf{18.09}} & \multicolumn{1}{l|}{\textbf{29.39}} & 0.328    \\
Thai              & \multicolumn{1}{l|}{19.94} & \multicolumn{1}{l|}{\textbf{27.63}} & \multicolumn{1}{c|}{\textbf{0.353}}    & \multicolumn{1}{l|}{\textbf{19.82}} & \multicolumn{1}{l|}{25.61} & 0.240    \\
Korean            & \multicolumn{1}{l|}{20.54} & \multicolumn{1}{l|}{27.63} & \multicolumn{1}{c|}{\textbf{0.338}}    & \multicolumn{1}{l|}{\textbf{18.69}} & \multicolumn{1}{l|}{\textbf{28.79}} & 0.284    \\
Hindi             & \multicolumn{1}{l|}{20.92} & \multicolumn{1}{l|}{25.90} & \multicolumn{1}{c|}{\textbf{0.338}}    & \multicolumn{1}{l|}{\textbf{18.52}} & \multicolumn{1}{l|}{\textbf{26.30}} & 0.311    \\
Ukrainian         & \multicolumn{1}{l|}{19.27} & \multicolumn{1}{l|}{28.18} & \multicolumn{1}{c|}{\textbf{0.346}}    & \multicolumn{1}{l|}{\textbf{17.36}} & \multicolumn{1}{l|}{\textbf{28.54}} & 0.314    \\
Arabic            & \multicolumn{1}{l|}{20.32} & \multicolumn{1}{l|}{28.71} & \multicolumn{1}{c|}{\textbf{0.346}}    & \multicolumn{1}{l|}{\textbf{18.34}} & \multicolumn{1}{l|}{\textbf{28.90}} & 0.298    \\
Turkey            & \multicolumn{1}{l|}{19.54} & \multicolumn{1}{l|}{28.53} & \multicolumn{1}{c|}{\textbf{0.347}}    & \multicolumn{1}{l|}{\textbf{17.40}} & \multicolumn{1}{l|}{\textbf{28.54}} & 0.315    \\
Vietnamese        & \multicolumn{1}{l|}{19.02} & \multicolumn{1}{l|}{29.22} & \multicolumn{1}{c|}{\textbf{0.346}}    & \multicolumn{1}{l|}{\textbf{17.02}} & \multicolumn{1}{l|}{\textbf{30.86}} & 0.312    \\
Polish            & \multicolumn{1}{l|}{19.67} & \multicolumn{1}{l|}{29.11} & \multicolumn{1}{c|}{\textbf{0.347}}    & \multicolumn{1}{l|}{\textbf{18.36}} & \multicolumn{1}{l|}{\textbf{30.41}} & 0.327    \\
Dutch             & \multicolumn{1}{l|}{20.14} & \multicolumn{1}{l|}{27.64} & \multicolumn{1}{c|}{\textbf{0.350}}    & \multicolumn{1}{l|}{\textbf{17.91}} & \multicolumn{1}{l|}{\textbf{29.48}} & 0.329    \\
Portuguese        & \multicolumn{1}{l|}{20.78} & \multicolumn{1}{l|}{\textbf{28.59}} & \multicolumn{1}{c|}{\textbf{0.352}}    & \multicolumn{1}{l|}{\textbf{18.82}} & \multicolumn{1}{l|}{28.56} & 0.302    \\
Italian           & \multicolumn{1}{l|}{19.77} & \multicolumn{1}{l|}{27.19} & \multicolumn{1}{c|}{\textbf{0.352}}    & \multicolumn{1}{l|}{\textbf{17.38}} & \multicolumn{1}{l|}{\textbf{28.53}} & 0.317    \\
Spanish           & \multicolumn{1}{l|}{\textbf{20.15}} & \multicolumn{1}{l|}{\textbf{27.64}} & \multicolumn{1}{c|}{\textbf{0.357}}    & \multicolumn{1}{l|}{20.41} & \multicolumn{1}{l|}{25.31} & 0.260    \\
German            & \multicolumn{1}{l|}{18.74} & \multicolumn{1}{l|}{27.58} & \multicolumn{1}{c|}{\textbf{0.359}}    & \multicolumn{1}{l|}{\textbf{17.06}} & \multicolumn{1}{l|}{\textbf{28.66}} & 0.347    \\
French            & \multicolumn{1}{l|}{18.99} & \multicolumn{1}{l|}{28.34} & \multicolumn{1}{c|}{\textbf{0.357}}    & \multicolumn{1}{l|}{\textbf{17.09}} & \multicolumn{1}{l|}{\textbf{29.71}} & 0.341    \\
Russian           & \multicolumn{1}{l|}{19.18} & \multicolumn{1}{l|}{28.26} & \multicolumn{1}{c|}{\textbf{0.347}}    & \multicolumn{1}{l|}{\textbf{17.49}} & \multicolumn{1}{l|}{\textbf{29.42}} & 0.322   \\ \hline
\end{tabular}
}
\caption{Comparison of zero-shot evaluation results with translation-based SD on MG-18.}
\label{Tab: translated_result}
\end{table}

We evaluate the general multilingual T2I generative capability of AD on MG-18. We compare AD with two kinds of models. The first is translation-based SD, which requires translating prompts in other languages into English before generation. The second is multilingual baseline diffusion models that beyond English. The inference resolution of all models is 512$\times$512, using 50 DDIM steps and 9.0 classifier-free guidance scale.

\noindent\textbf{Metrics} We use FID and IS for evaluating the generation quality, and use Multilingual CLIP\footnote{https://huggingface.co/laion/CLIP-ViT-H-14-frozen-xlm-roberta-large-laion5B-s13B-b90k} to calculate the cosine similarity score(CLIP Sim) to evaluate the consistency of generated images with multilingual text. 

\noindent\textbf{Compare with Translation-based SD v2.1} We use the original multilingual prompts directly as the input of AltDiffuison. Considering that SD only supports English inputs, we first use the state-of-the-art opensource translation model NLLB-3B\footnote{https://huggingface.co/facebook/nllb-200-3.3B} to translate other languages into English and then feed them into SD. 

As shown in Table~\ref{Tab: translated_result}, AD surpasses translation-based SD in CLIP Sim of all languages and achieves comparable results in FID and IS, proving that AD is better than SD in multilingual understanding and can generate almost the same high-quality images as SD on general prompts.



\begin{table}[htbp]
\centering
\resizebox{\columnwidth}{!}{
\begin{tabular}{c|c|c|cl}
\hline
Language & Model                     & FID(↓) & IS(↑) & CLIP Sim(↑) \\ \hline
English & Taiyi-Bilingual        & 25.76  & 26.54 & 0.261       \\
& AltDiffusion(AD) & \textbf{19.02}  & \textbf{27.45} & \textbf{0.324$_{(\textcolor[rgb]{0.13,0.55,0.13}{+24.1\%})}$}  \\ \hline
Chinese & Taiyi-CN               & 20.72  & 28.91 & 0.276      \\
& Taiyi-Bilingual       & 23.87  & 26.96 & 0.259         \\ 
& AltDiffusion(AD)       & \textbf{20.32}  & \textbf{29.46} & \textbf{0.350$_{(\textcolor[rgb]{0.13,0.55,0.13}{+26.8\%})}$}         \\ \hline
Japanese & Japanese Stable Diffusion(SD) & 22.78  & 30.57 & 0.278         \\
& AltDiffusion(AD)   & \textbf{18.90}  & \textbf{28.13} & \textbf{0.356$_{(\textcolor[rgb]{0.13,0.55,0.13}{+28.1\%})}$}         \\ \hline
\end{tabular}
}
\caption{Comparison of zero-shot evaluation results with other multilingual baselines on MG-18.}
\label{Tab: compared_result}
\end{table}

\noindent\textbf{Compare with Other Baselines} We compare AD with other multilingual baseline diffusion models, including Taiyi Chinese, Taiyi Bilingual ,and Japanese SD.

As shown in Table~\ref{Tab: compared_result}, AD outperforms other multilingual baseline diffusion models in all metrics, especially in CLIP Sim, AD has achieved 24.1\%, 26.8\% and 28.1\% percent improvement on English, Chinese and Japanese respectively. It shows that AD has stronger image generation capability and multilingual understanding capability than other multilingual baseline models.

To demonstrate the strong capability of AD in multilingual T2I generation, we provide generated images in various languages in Figure~\ref{fig: generated_exmaples}. More generated results can be found in the Appendix. As shown in Figure~\ref{fig: compared_results.pdf}(b) and (c), AD can generate images that are more consistent with multilingual text, while other models often ignore or make a mistake in concepts. For example, ``black couch'', ``wood floor'' and ``parasol'' in Chinese, and ``bears'' and ``yellow bus'' in Japanese.
AD can generate results comparable to SD and better than Taiyi Bilingual in English, as shown in Figure~\ref{fig: compared_results.pdf}(a).


\subsection{Results on MC-18}
\begin{figure}[htbp] 
\centering 
\includegraphics[width=1\columnwidth]{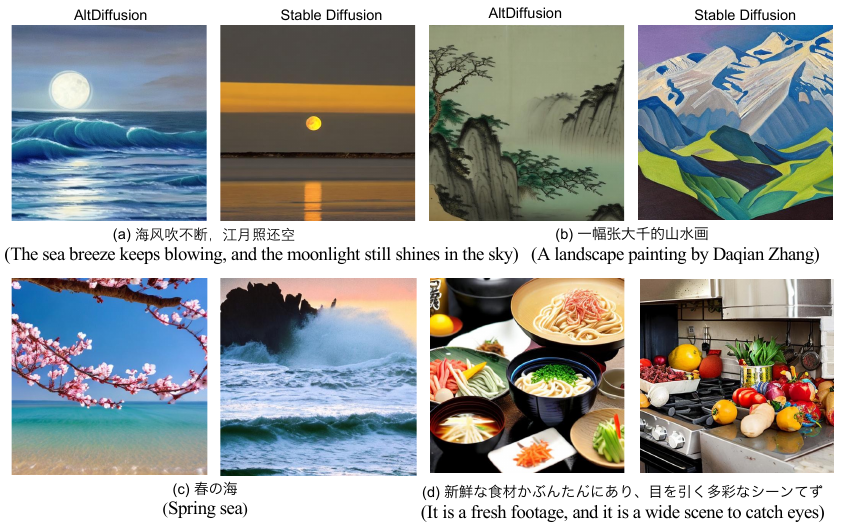} 
\caption{Comparison of generated results with translation-based SD on MC-18. AD generates images that are more appropriate to culture-specific concepts.}
\label{fig: visualization_mc_18}
\end{figure}

\begin{figure*}[htbp] 
\centering 
\includegraphics[width=0.95 \textwidth]{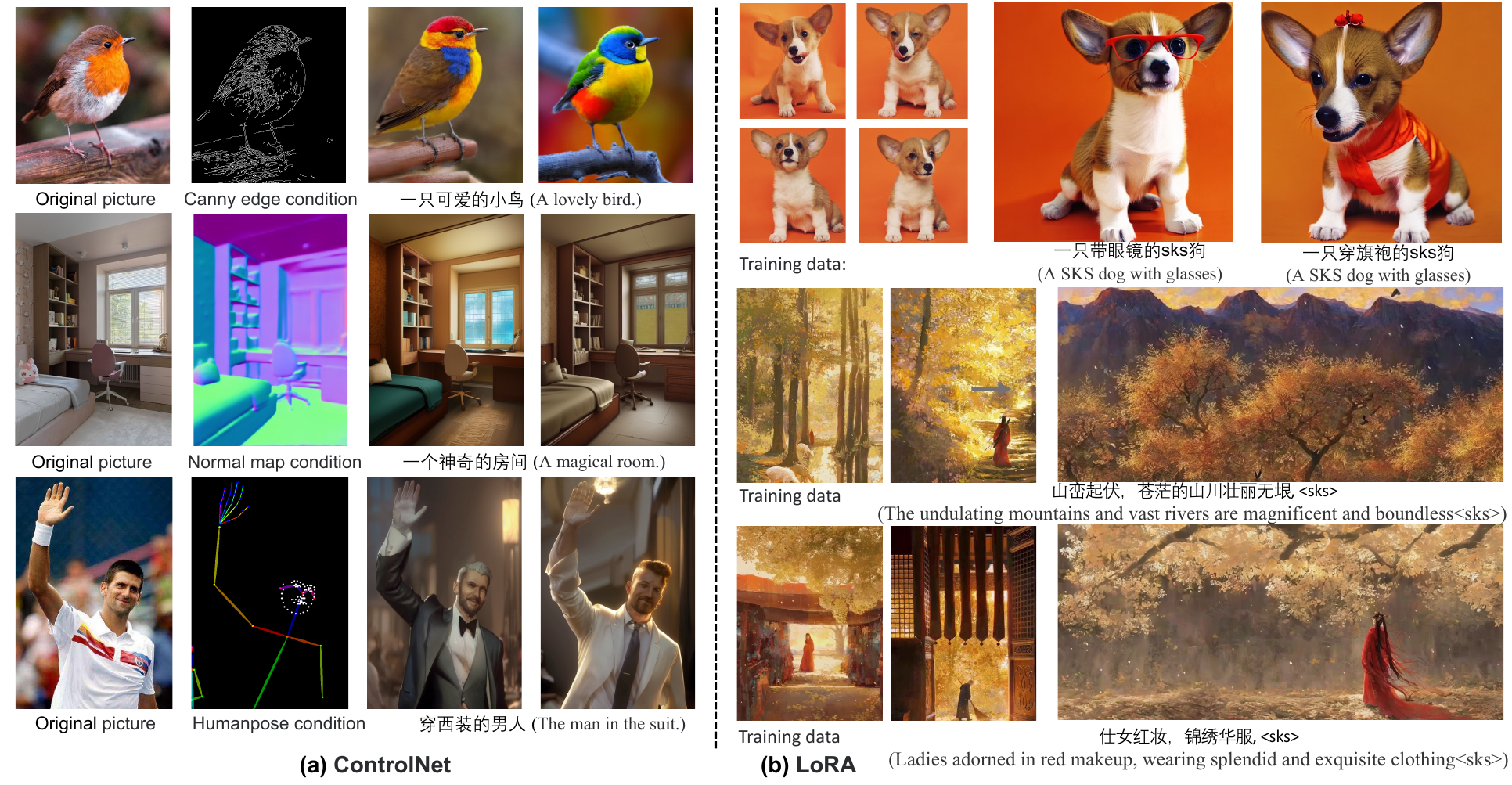} 
\caption{AD has strong compatibility with downstream T2I tools such as ControlNet and LoRA. }
\label{fig: aplication2}
\end{figure*}
Annotators familiar with culture of different countries are asked to conduct a human evaluation for the understanding capability of the model in culture-specific concepts.

\noindent \textbf{Evaluation Setting} We assign three annotators to each language. Annotators see two images generated by the same prompt generated by AD and SD, respectively. The evaluation interface can be viewed in the Appendix. Then they are asked to score from two dimensions: Culture Consistency, and Image-Text Consistency, scoring from 1-5. After scoring, the annotators select the final result from [``Alt is better'', ``SD is better'', ``Same'']. Then we calculate the total score according to the following formula:
\begin{scriptsize}
\begin{equation}
    \left(Total_{Alt}, Total_{SD}\right) = \left(\frac{\left|A \right|+0.5\cdot\left|C \right|}{N}, \frac{\left|B \right|+0.5\cdot\left|C \right|}{N}\right)
\end{equation}
\end{scriptsize}

\noindent where $\left|A \right|, \left|B \right| and \left|B \right|$ is the count of ``Alt is better'',``SD is better'' and ``Same'', $N=\left|A \right|+\left|B \right|+\left|C \right|$. The results for each language are the average of 3 annotator scores.

\noindent \textbf{Evaluation Results} As shown in Table~\ref{Tab: human_evaluation_result}, AD beats SD for the final total scores in all languages and outstands in Cultural and Image-Text Consistency, showing that AD performs better in multilingual understanding. The evaluation results indicate that through training with large-scale multilingual data, culture-specific concepts of different languages can be injected into the model.

\begin{table}[htbp]
\centering
\resizebox{\columnwidth}{!}{
\begin{tabular}{c|ccc|ccc}
\hline
\multirow{3}{*}{} & \multicolumn{3}{c|}{AltDiffusion(AD)} & \multicolumn{3}{c}{Stable Diffusion(SD)} \\ \cline{2-7} 
                  & Culture & Text-Image & \multirow{2}{*}{Total} & Culture & Text-Image & \multirow{2}{*}{Total} \\ 
                  & Consistency & Consistency & & Consistency & Consistency & \\ \hline
Chinese           & \multicolumn{1}{c|}{\textbf{4.125}}         & \multicolumn{1}{c|}{\textbf{3.775}}  & \textbf{0.769} & \multicolumn{1}{c|}{3.356}         & \multicolumn{1}{c|}{3.350}   & 0.231 \\
Japanese          & \multicolumn{1}{c|}{\textbf{4.507}}         & \multicolumn{1}{c|}{\textbf{4.487}}   & \textbf{0.757} & \multicolumn{1}{c|}{3.301}         & \multicolumn{1}{c|}{3.253}   & 0.243 \\
Thai              & \multicolumn{1}{c|}{\textbf{4.027}}         & \multicolumn{1}{c|}{\textbf{4.044}}   & \textbf{0.648} & \multicolumn{1}{c|}{3.383}         & \multicolumn{1}{c|}{3.579}   & 0.352 \\
Korean            & \multicolumn{1}{c|}{\textbf{3.236}}         & \multicolumn{1}{c|}{\textbf{3.371}}   & \textbf{0.607} & \multicolumn{1}{c|}{3.135}         & \multicolumn{1}{c|}{3.287}   & 0.393 \\
Hindi             & \multicolumn{1}{c|}{\textbf{4.824}}         & \multicolumn{1}{c|}{\textbf{4.301}}   & \textbf{0.523} & \multicolumn{1}{c|}{4.784}         & \multicolumn{1}{c|}{4.261}   & 0.477 \\
Ukrainian         & \multicolumn{1}{c|}{\textbf{4.422}}         & \multicolumn{1}{c|}{\textbf{3.955}}   & \textbf{0.641} & \multicolumn{1}{c|}{4.322}         & \multicolumn{1}{c|}{3.905}   & 0.359 \\
Arabic            & \multicolumn{1}{c|}{\textbf{4.824}}         & \multicolumn{1}{c|}{\textbf{4.401}}   & \textbf{0.609} & \multicolumn{1}{c|}{4.647}         & \multicolumn{1}{c|}{4.314}   & 0.391 \\
Turkey            & \multicolumn{1}{c|}{\textbf{3.376}}         & \multicolumn{1}{c|}{\textbf{3.293}}   & \textbf{0.510} & \multicolumn{1}{c|}{3.328}         & \multicolumn{1}{c|}{3.241}   & 0.490 \\
Vietnamese        & \multicolumn{1}{c|}{\textbf{3.637}}         & \multicolumn{1}{c|}{\textbf{3.511}}   & \textbf{0.533} & \multicolumn{1}{c|}{3.467}         & \multicolumn{1}{c|}{3.396}   & 0.467 \\
Polish            & \multicolumn{1}{c|}{\textbf{4.243}}         & \multicolumn{1}{c|}{\textbf{3.735}}   & \textbf{0.546} & \multicolumn{1}{c|}{4.130}         & \multicolumn{1}{c|}{3.676}   & 0.454 \\
Dutch             & \multicolumn{1}{c|}{\textbf{4.495}}         & \multicolumn{1}{c|}{\textbf{4.465}}   & \textbf{0.527} & \multicolumn{1}{c|}{4.195}         & \multicolumn{1}{c|}{4.215}   & 0.473 \\
Portuguese        & \multicolumn{1}{c|}{\textbf{3.757}}         & \multicolumn{1}{c|}{\textbf{3.627}}   & \textbf{0.639} & \multicolumn{1}{c|}{3.639}         & \multicolumn{1}{c|}{3.509}   & 0.361 \\
Italian           & \multicolumn{1}{c|}{\textbf{3.586}}         & \multicolumn{1}{c|}{\textbf{3.449}}   & \textbf{0.565} & \multicolumn{1}{c|}{3.512}         & \multicolumn{1}{c|}{3.375}   & 0.435 \\
Spanish           & \multicolumn{1}{c|}{\textbf{4.202}}         & \multicolumn{1}{c|}{\textbf{4.113}}   & \textbf{0.554} & \multicolumn{1}{c|}{4.138}         & \multicolumn{1}{c|}{4.064}   & 0.446 \\
German            & \multicolumn{1}{c|}{\textbf{3.981}}         & \multicolumn{1}{c|}{\textbf{3.916}}   & \textbf{0.698} & \multicolumn{1}{c|}{3.825}         & \multicolumn{1}{c|}{3.688}   & 0.302 \\
French            & \multicolumn{1}{c|}{\textbf{4.655}}        & \multicolumn{1}{c|}{\textbf{4.582}}   & \textbf{0.527}  & \multicolumn{1}{c|}{4.652}        & \multicolumn{1}{c|}{4.579}  & 0.473 \\
Russian           & \multicolumn{1}{c|}{\textbf{3.258}}         & \multicolumn{1}{c|}{\textbf{3.086}}   & \textbf{0.556} & \multicolumn{1}{c|}{2.868}         & \multicolumn{1}{c|}{3.002}   & 0.444 \\ \hline
\end{tabular}
}
\caption{Comparison of human evaluation results with translation-based SD on MC-18.}
\label{Tab: human_evaluation_result}
\end{table}

We illustrate the performance difference between AD and SD on MC-18 in Figure~\ref{fig: visualization_mc_18}. AD has a better understanding capability in culture-specific concepts. For example, in Figure~\ref{fig: visualization_mc_18}(a), the prompt is actually a Chinese poem, but SD generates a realistic image. In Figure~\ref{fig: visualization_mc_18}(b), models are asked to generate a landscape painting by Chinese traditional painter Daqian Zhang. The image of AD shows the characteristics of Chinese traditional painting, while the image of SD is an oil painting. As for ``the spring sea'' in Japanese in In Figure~\ref{fig: visualization_mc_18}(c), AD generates cherry blossoms more suitable for artistic concept. 


\section{Application}

\begin{figure}[htbp] 
\centering 
\includegraphics[width=0.95\linewidth]{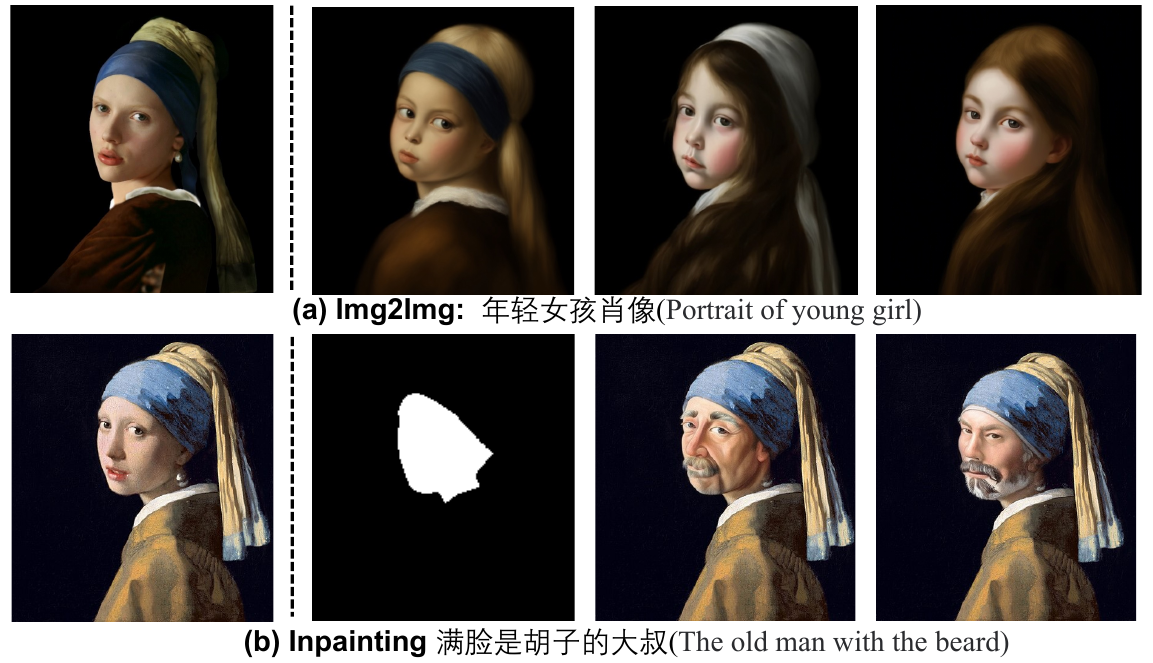} 
\caption{The images generated by Img2Img and Inpainting function using AD .}
\label{fig: application_1}
\end{figure}

\noindent \textbf{Genernal Application} Figure~\ref{fig: application_1} shows the general capacities of AD, such as Image to Image(Img2Img) and Inpainting. AD supports users to to use Image to Image or Inpaint function directly use languages beyond English, e.g., Chinese.

\begin{figure}[htbp] 
\centering 
\includegraphics[width=0.95\columnwidth]{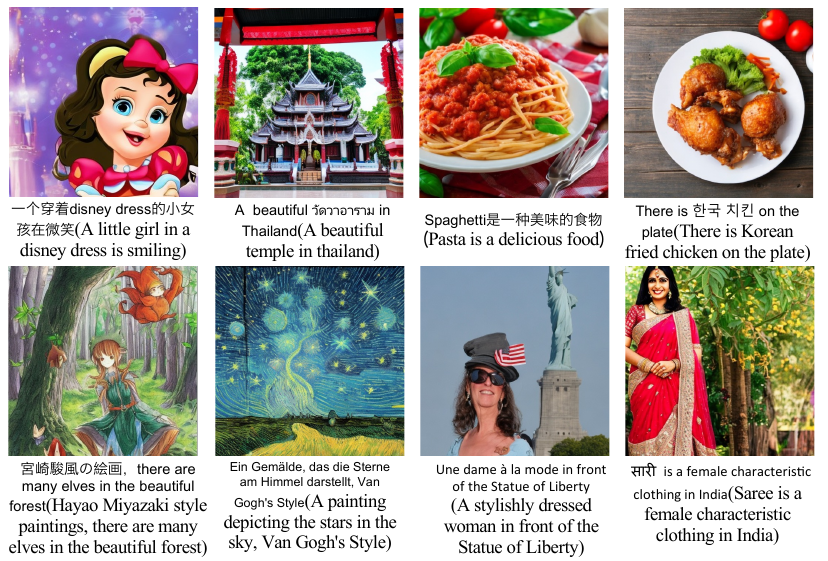} 
\caption{Images generated by mixed languages using AD.}
\label{fig: mixed_languages}
\end{figure}

\noindent \textbf{Compatibility with Downstream Tools} Although large T2I models achieve impressive performance, they still lack controllability in specific commercial applications. Recently, some methods, such as ControlNet\cite{controlnet} and LoRA\cite{lora}, have garnered widespread attention for enhancing model controllability. Compatibility with these downstream tools is essential for a Large T2I. As shown in Figure~\ref{fig: aplication2}, AD is totally compatible with ControlNet and LoRA. Thus users can use their imagination to create images easily.


\noindent \textbf{Mixed Language Generation} As shown in Figure~\ref{fig: mixed_languages}, AD supports mixed language input. It will be very troublesome if the model can only support English because users need to translate various languages into English and then concatenate them as input. AD can freely combine different languages, such as Thai and English, Japanese and English, Chinese and Korean, etc.

\section{Conclusion}
This paper introduces AltDiffusion(AD), a multilingual T2I diffusion model that supports eighteen languages. We train a multilingual text encoder and plug it into pretrained diffusion model, and then train the diffusion model using a two-stage training schema. In addition, we introduce a benchmark to evaluate AD, including two datasets focusing on general and culture-specific evaluation: MG-18 and MC-18. Experimental results show AltDiffusion outperforms current state-of-the-art T2I models, e.g., Stable Diffusion in multilingual understanding, especially with respect to culture-specific concepts, while still having comparable capability for generating high-quality images. Meanwhile, as a large multilingual T2I diffusion model, AD is compatible with all downstream T2I tools, e.g., ControlNet and LoRA, which may promote research and application in multilingual T2I.

\bibliography{aaai24}

\clearpage

\section{Appendix}

\subsection{A. Hyperparameters \label{a.1}}


\begin{figure}[ht] 
\centering 
\includegraphics[width=1.0\columnwidth]{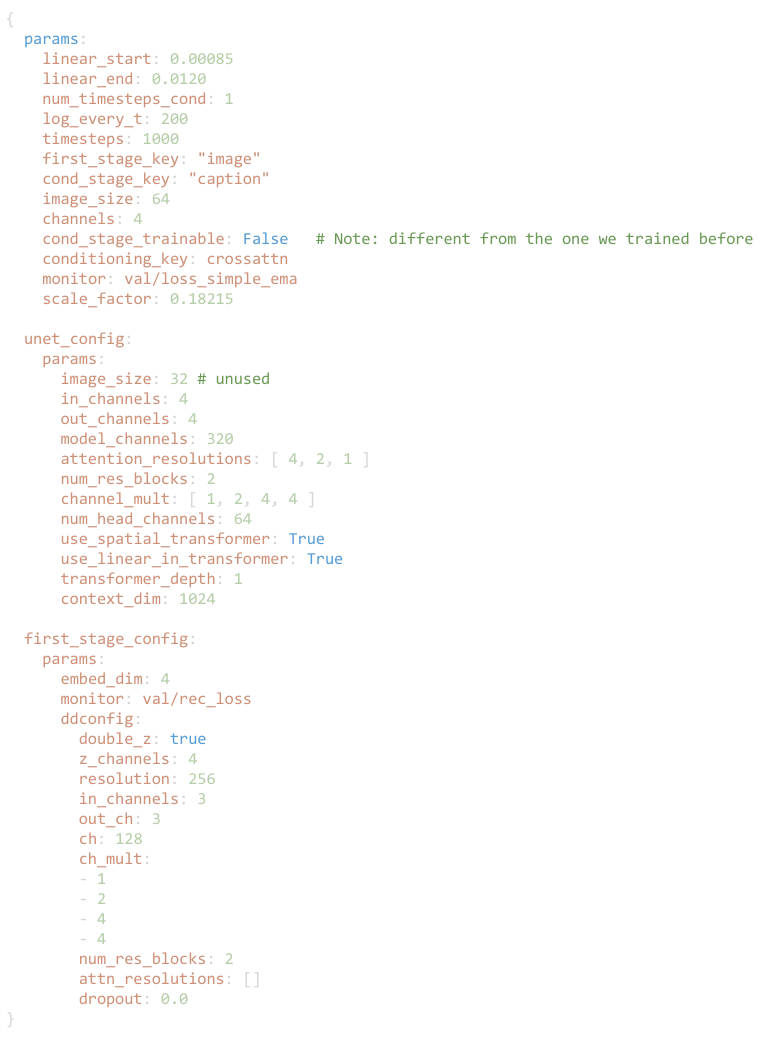} 
\caption{Hyperparameter}
\label{fig: hyperparameter}
\end{figure}

\subsection{B. Reuslts on MMS-COCO}
\begin{table}[htbp]
\centering
\resizebox{\columnwidth}{!}{
\begin{tabular}{c|c|ccc}
\hline
Language & Model                     & FID(↓) & IS(↑) & CLIP score(↑) \\ \hline
English & Stable Diffusion v2.1     & 22.53  & 35.66 & 0.337         \\
& Taiyi-Bilingual        & 33.17  & 31.50 & 0.255         \\
& \textbf{AltDiffusion}       & \textbf{23.94}  & \textbf{35.37} & \textbf{0.335 }  \\ \hline
Chinese & Taiyi-CN               & 25.23  & 33.36 & 0.265      \\
& Taiyi-Bilingual       & 26.84  & 33.24 & 0.256         \\ 
& \textbf{AltDiffusion}       & \textbf{24.02}  & \textbf{33.78} & \textbf{0.319}          \\ \hline
Japanese & Japanese Stable Diffusion & 28.70  & 32.72 & 0.276         \\
& \textbf{AltDiffusion}       & \textbf{23.94}  & \textbf{33.53} & \textbf{0.326 }         \\ \hline
Thai & & 24.22  & 33.48 & 0.313            \\ 
Korean & & 24.43  & 33.66 & 0.318             \\ 
Hindi & & 23.99  & 33.49 & 0.321              \\ 
Ukrainian & & 24.50  & 33.33 & 0.320              \\ 
Arabic & & 24.07  & 34.46 & 0.315              \\ 
Turkey & & 23.78  & 33.89 & 0.313              \\ 
Vietnamese & AltDiffusion & 24.15  & 34.09 & 0.318              \\ 
Polish & & 24.22  & 34.78 & 0.323              \\ 
Dutch & & 24.17  & 33.93 & 0.326              \\ 
Portuguese & & 24.37  & 33.57 & 0.325              \\ 
Italian & & 24.44  & 33.61 & 0.323              \\ 
Spanish & & 24.11  & 34.33 & 0.312              \\ 
German & & 23.92  & 34.25 & 0.328              \\ 
French & & 24.32  & 33.99 & 0.325              \\ 
Russian & & 24.56  & 34.18 & 0.320              \\ \hline
\end{tabular}
}
\caption{Zero-shot evaluation results on MMS-COCO(10K from validation dataset) with classifier-free guidance scale 9.0.}
\label{Tab: mmscoco_result}
\end{table}
We build multilingual MS-COCO(MMS-COCO) by random sample 10000 prompts from the MS-COCO 2017 validation dataset\cite{mscoco} and translate them into other 17 languages by using mBart\cite{mbart}. We evaluate zero-shot performance of AltDiffusion on MMS-COCO, results is shown on Table~\ref{Tab: mmscoco_result}.

\subsection{C. Classifier guidance scale influence}

\begin{figure}[htbp] 
\centering 
\includegraphics[width=1.\columnwidth]{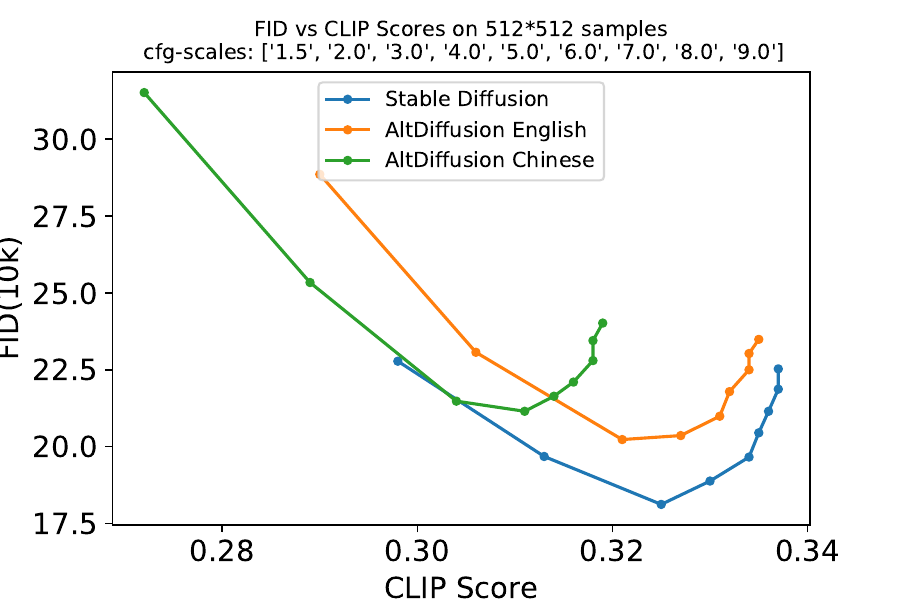} 
\caption{Influence of cfg scale.}
\label{fig: cfg}
\end{figure}

Figure~\ref{fig: cfg} shows the influence of classifier guidance scale influence on MMS-COCAO in AltDiffusion-Chinese, AltDiffusion-English and Stable Diffusion. Classifier guidance scale from 1.5 to 9.0.

\subsection{D. Human Evaluation Interface}

\begin{figure*}[htbp] 
\centering 
\includegraphics[width=0.90\textwidth]{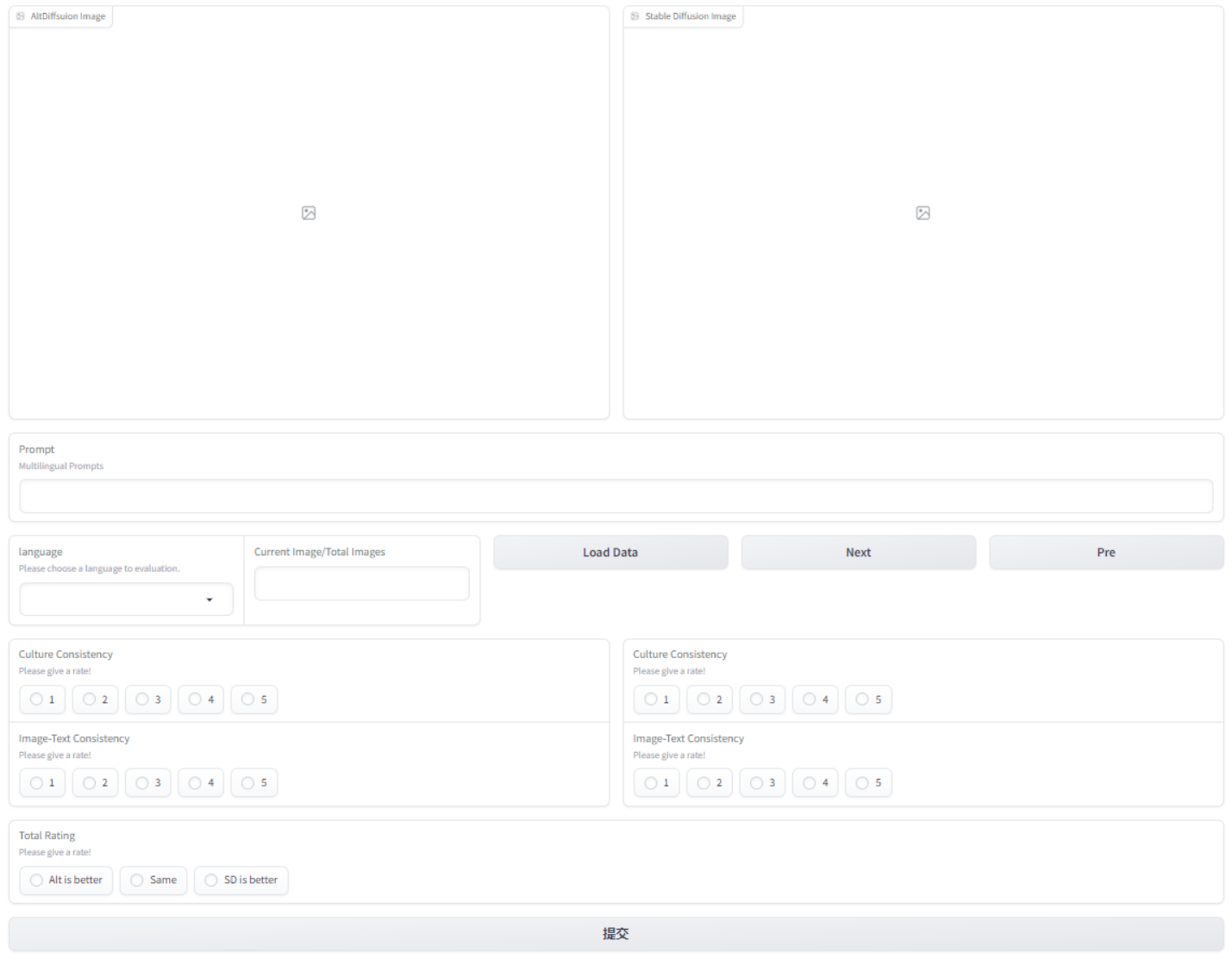} 
\caption{Human Evaluation Interface.}
\label{fig: interface}
\end{figure*}

We design the human evaluation interface as shown in Figure~\ref{fig: interface}.

\subsection{E. Training Stage Abalation}

\begin{figure*}[htbp] 
\centering 
\includegraphics[width=0.90\textwidth]{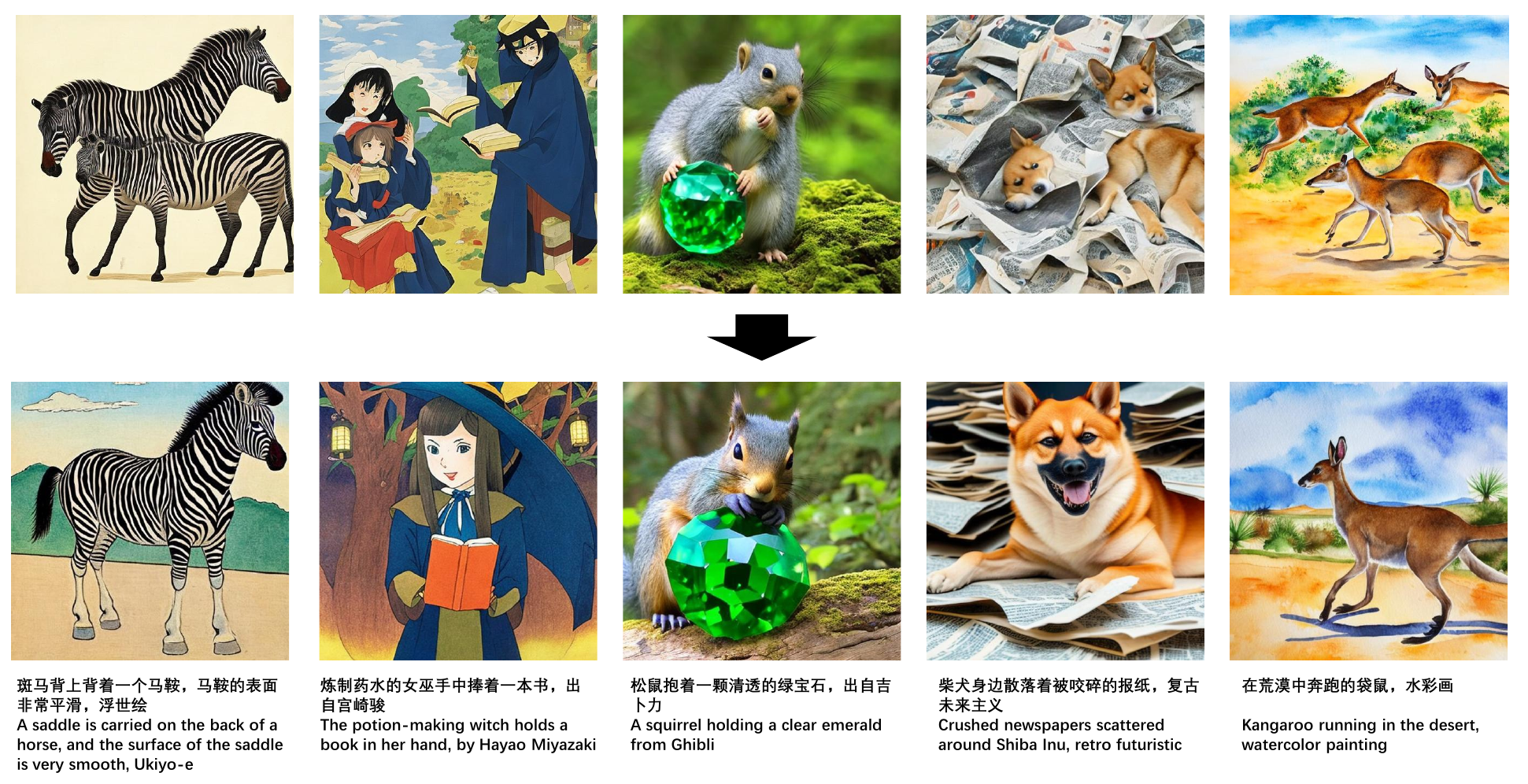} 
\caption{Comparing the generated result of Stage one and two. The first row is the generated result of the first stage. The second row is the generated result of the second stage. The second stage of training improves the quality of generated images.}
\label{fig: stage_compare.pdf}
\end{figure*}

We demonstrate the effect of our two-stage training method in Figure~\ref{fig: stage_compare.pdf}. The first row is the generated result of the first stage and the second row is the generated result of the second stage. After the first stage of training, images matching text can be generated and the generation quality can be futher improved in the second stage of training. We can see that the second stage is still very necessary in improving the quality of generated images, which can effectively avoid discordant and repetitive phenomena.

\subsection{F. More Generated Results}

\begin{figure*}[htbp]
  \centering 
  \includegraphics[width=0.9\textwidth]{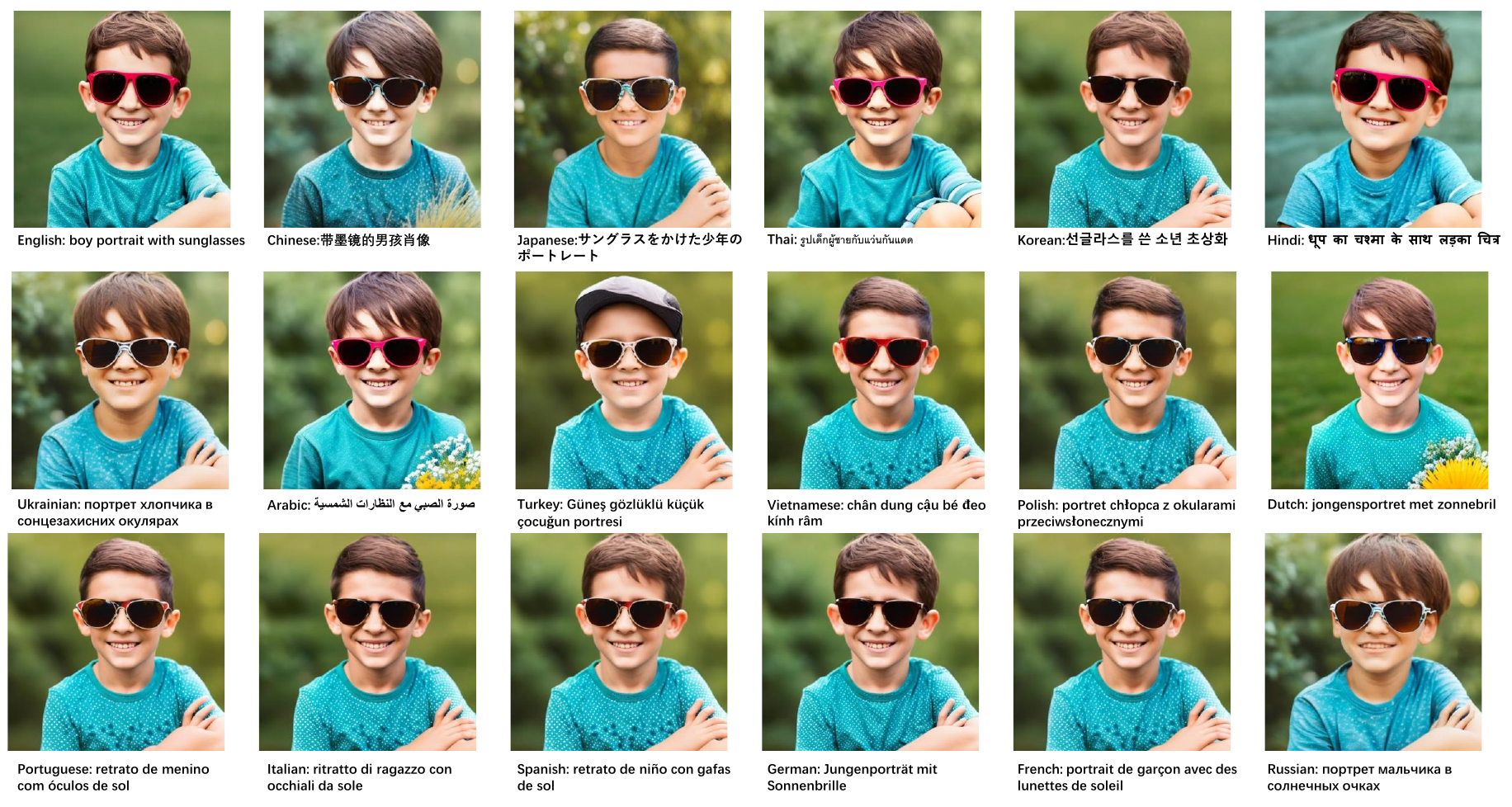}
  \caption{ Images generated by AltDiffusion with prompt ``boy portrait with sunglasses'' in various languages and a fixed seed. Note that the model demonstrates proficiency in capturing distinct facial features of young males from various cultural backgrounds, including a European-American style for English and an Asian style for Chinese.}
  \label{fig: boy}
\end{figure*}

\begin{figure*}[htbp] 
\centering 
\includegraphics[width=0.85\textwidth]{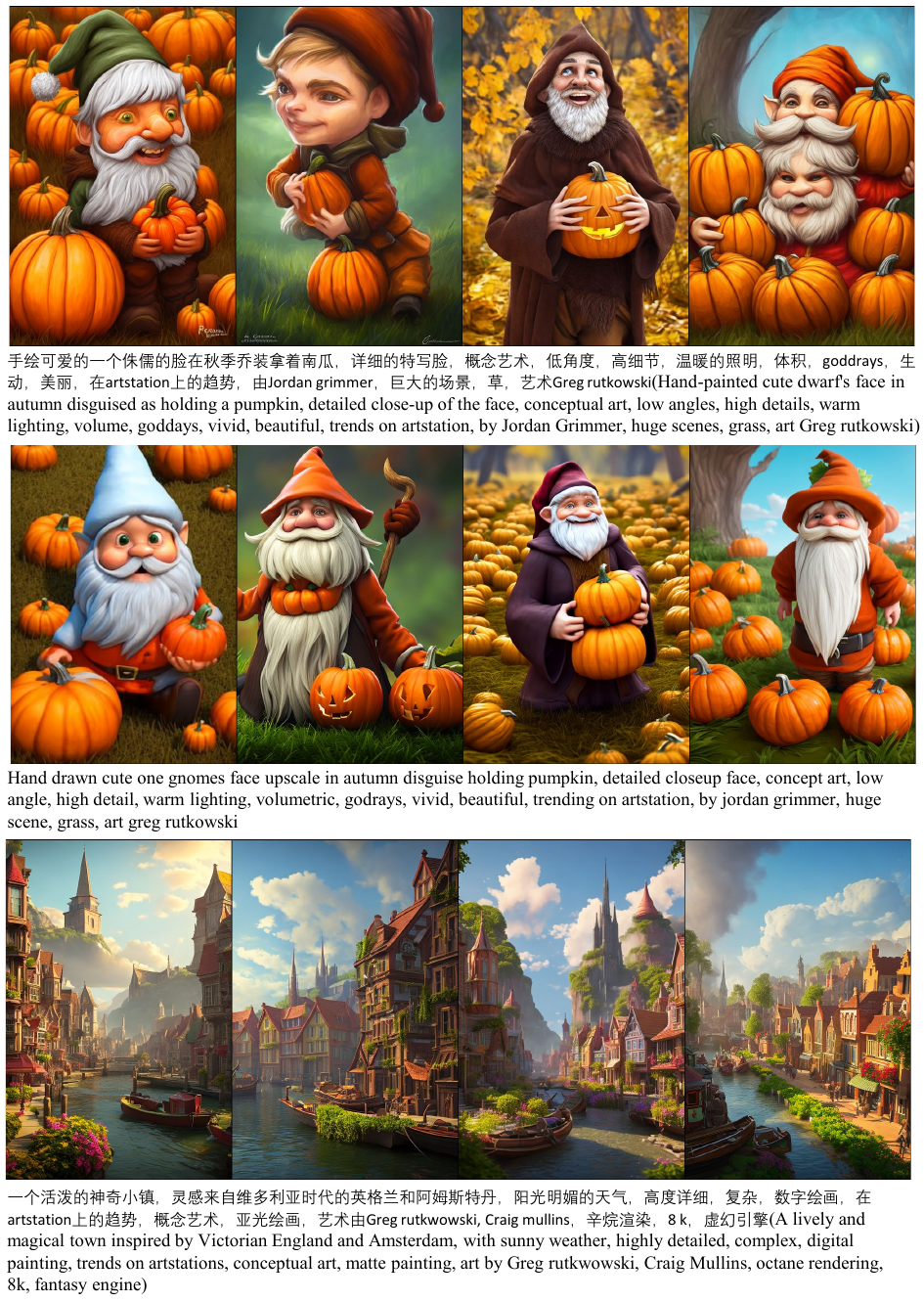} 
\caption{Generated images of in long format of 512*768 resolution.}
\label{fig: long1}
\end{figure*}

\begin{figure*}[htbp] 
\centering 
\includegraphics[width=0.85\textwidth]{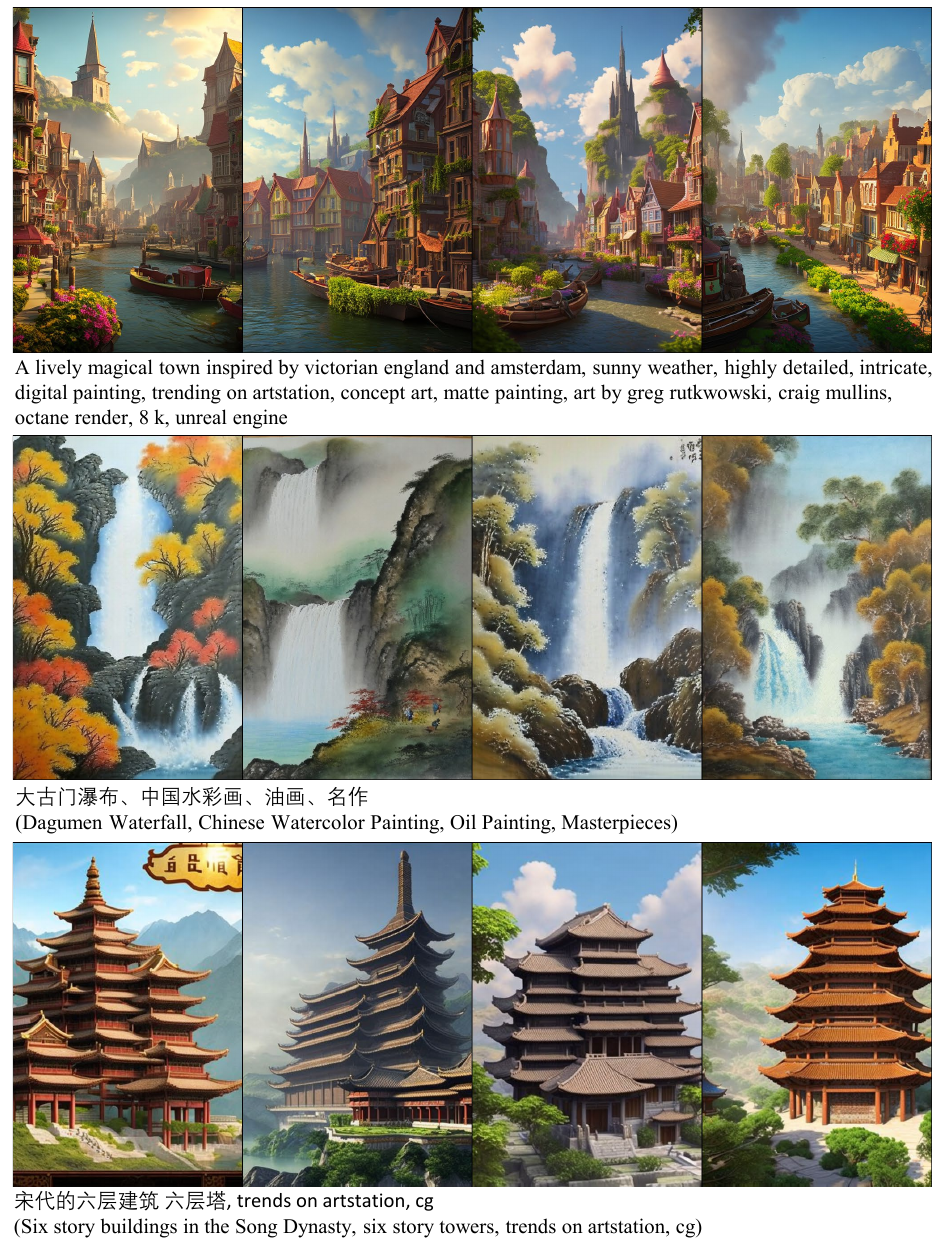} 
\caption{Generated images of in long format of 512*768 resolution.}
\label{fig: long2}
\end{figure*}

\begin{figure*}[htbp] 
\centering 
\includegraphics[width=0.85\textwidth]{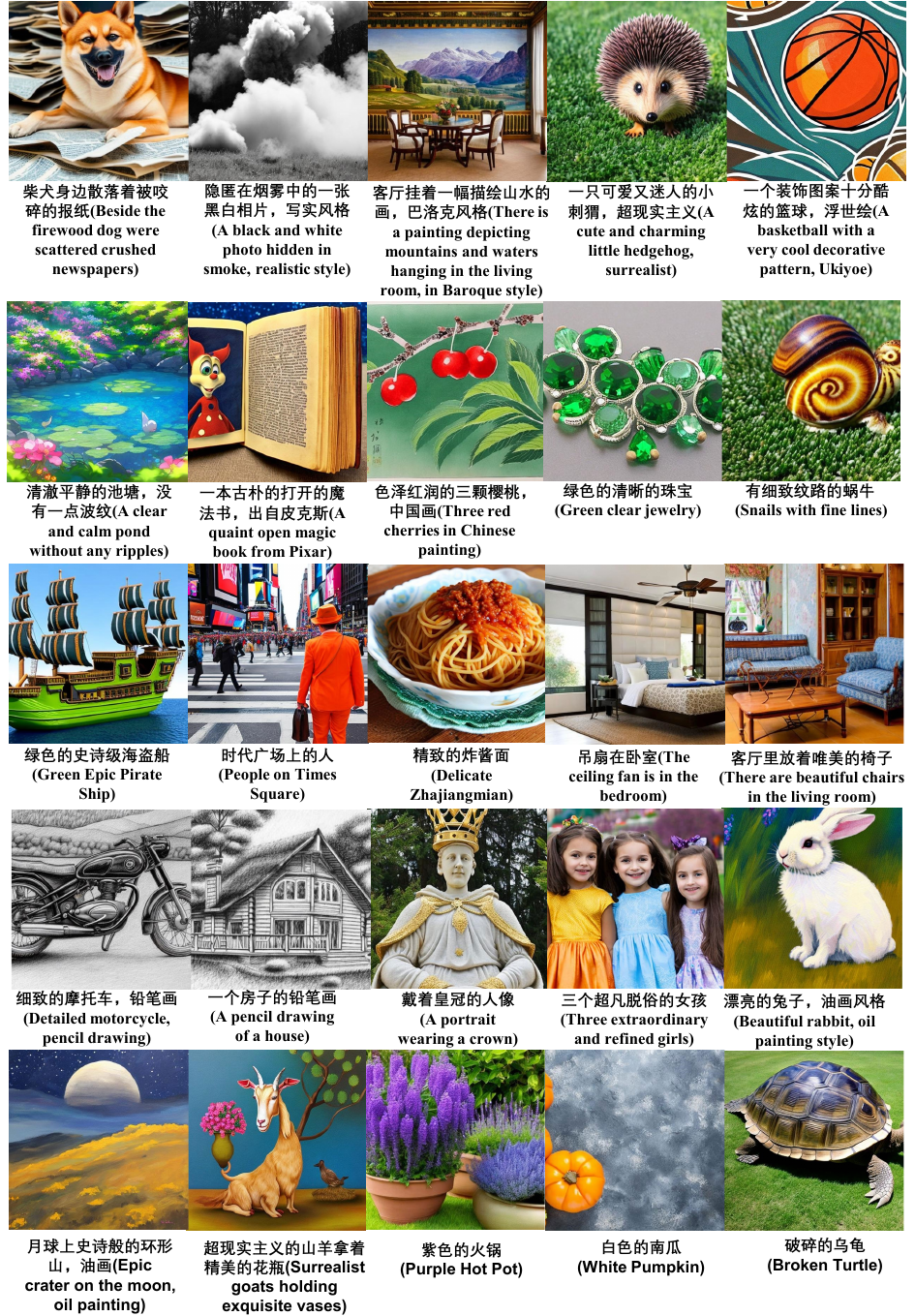} 
\caption{Uncurated examples of Chinese prompts. AltDiffusion works well for the most part, with the occasional base case that generates the wrong content.}
\label{fig: chinese_samples}
\end{figure*}

We show more generated results of AltDiffusion. Figure~\ref{fig: boy} and Figure~\ref{fig: boy} shown the images be generated using same prompt but in different languages. Figure~\ref{fig: long1} and Figure~\ref{fig: long2} shown the ability of AltDiffusion to generate larger images. Figure~\ref{fig: chinese_samples} shows uncurated examples in Chinese prompts.

\begin{figure*}[htbp] 
\centering 
\includegraphics[width=0.96\textwidth]{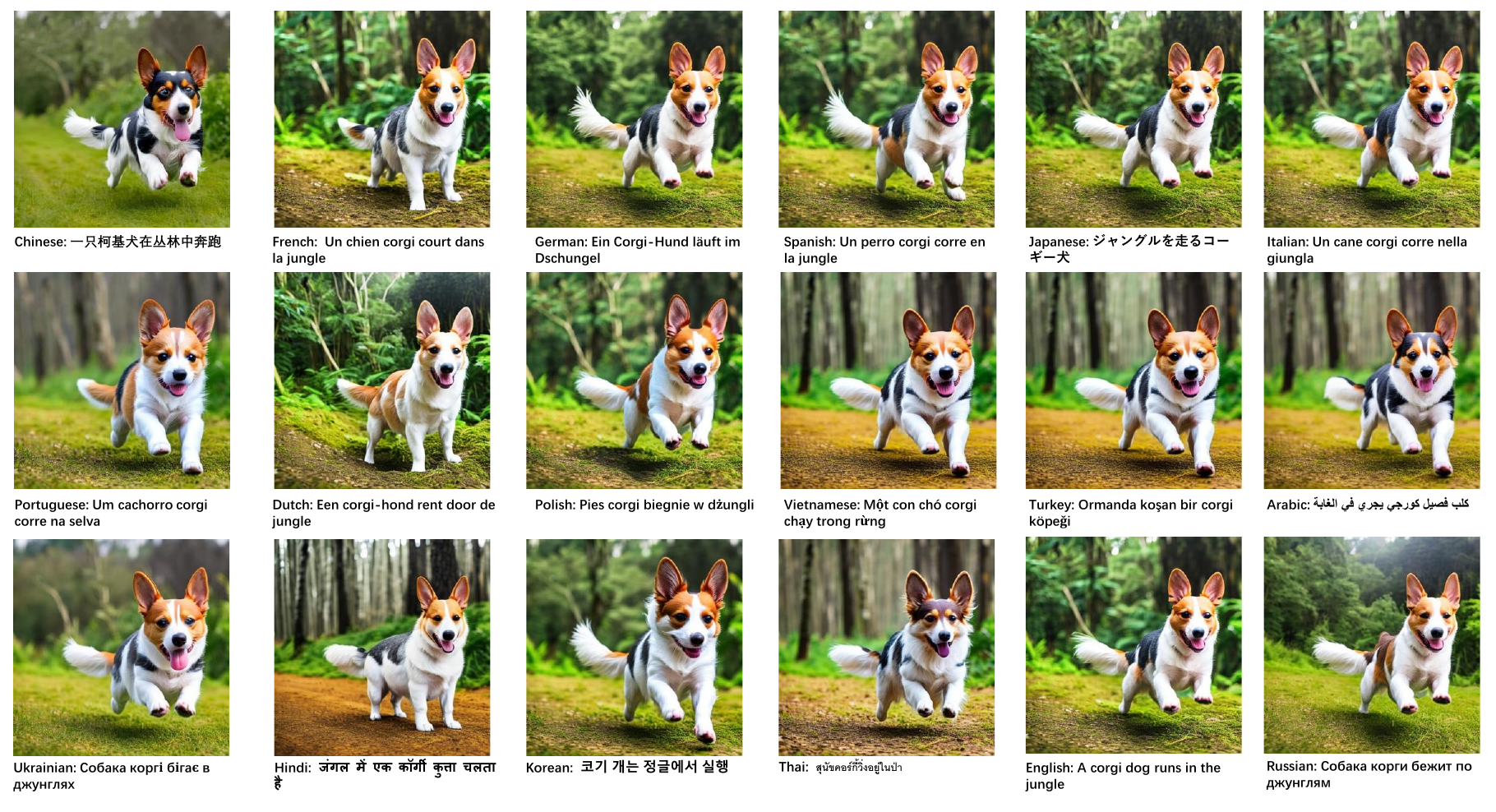} 
\caption{Generated images of ``A corgi dog runs in the jungle'' in different languages.}
\label{fig: dog}
\end{figure*}

\end{document}